# Hi-BEHRT: Hierarchical Transformer-based model for accurate prediction of clinical events using multimodal longitudinal electronic health records

Yikuan Li, Mohammad Mamouei, Gholamreza Salimi-Khorshidi, Shishir Rao, Abdelaali Hassaine, Dexter Canoy, Thomas Lukasiewicz, and Kazem Rahimi

***Abstract*— Electronic health records represent a holistic overview of patients' trajectories. Their increasing availability has fueled new hopes to leverage them and develop accurate risk prediction models for a wide range of diseases. Given the complex interrelationships of medical records and patient outcomes, deep learning models have shown clear merits in achieving this goal. However, a key limitation of these models remains their capacity in processing long sequences. Capturing the whole history of medical encounters is expected to lead to more accurate predictions, but the inclusion of records collected for decades and from multiple resources can inevitably exceed the receptive field of the existing deep learning architectures. This can result in missing crucial, long-term dependencies. To address this gap, we present Hi-BEHRT, a hierarchical Transformer-based model that can significantly expand the receptive field of Transformers and extract associations from much longer sequences. Using a multimodal large-scale linked longitudinal electronic health records, the Hi-BEHRT exceeds the state-of-the-art BEHRT 1% to 5% for area under the receiver operating characteristic (AUROC) curve and 3% to 6% for area under the precision recall (AUPRC) curve on average, and 3% to 6% (AUROC) and 3% to 11% (AUPRC) for patients with long medical history for 5-year heart failure, diabetes, chronic kidney disease, and stroke risk prediction. Additionally, because pretraining for hierarchical Transformer is not well-established, we provide an effective end-to-end contrastive pre-training strategy for Hi-BEHRT using EHR, improving its transferability on predicting clinical events with relatively small training dataset.**

## I. Introduction

Risk models play an important role in disease prognosis, diagnosis, and intervention. Currently, most risk models are conventional statistical models based on expert-selected predictors. For instance, QRISK[1], Framingham risk score[2], and ASSIGN risk score[3] are commonly used models for cardiovascular diseases. However, with the growing access to electronic health records (EHR), especially linked longitudinal EHR, from millions of patients, we now have an unprecedented opportunity to achieve a better understanding of patients' health trajectories, and to develop novel risk prediction models, which can capture important predictors and their long-term interdependencies towards more accurate risk prediction.

EHR provide up-to-date and comprehensive information about patients. It allows a clinician to assess the entire patient journey and represents what is actually available in clinical practice[4]. Due to its complexity and heterogeneity, modelling using large-scale EHR is a challenge. Some of the previous works have shown that deep learning is an effective method and deep learning models outperform standard statistical models in various complex risk prediction tasks[5]–[7] using EHR. However, existing works as applied to EHR have largely been based on US EHR[8] and relied on a fraction of information available in the datasets (typically disease and medications from hospital records). This would typically involve maximally a few hundred records from a patient[5], [7], [9]. A realistic demand of using more comprehensive information from the records can substantially increase the number of available records and expand the EHR sequence length for modelling. Therefore, adapting risk prediction model to handle patients with thousands of records or even longer EHR sequence and avoid missing important historical information is highly desired, but still remains as a critical bottleneck. For instance, Transformer-based models have gained wide popularity for risk prediction using EHR due to their superior performance for handling sequential data[9], [10]. However, their complexity grows quadratically as the sequence length grows[11], and a Transformer model with over 100 million parameters can only handle a sequence with maximally 512 sequence length[12].

Recent research has proposed two potential solutions for the Transformer-based models: (1) the use of sparse attention to replace the classic self-attention mechanism in Transformer[13], [14]; (2) and the use of hierarchical model architectures to extract local temporal features and reduce sequence length before feeding into the higher-level Transformer architecture[15]. Both approaches substantially reduce the complexity of Transformer models and adapt them to better handle data with longer sequence. Considering that medical events naturally have stronger local correlation (i.e., the closer two events are in time, the more likely that they are also semantically related), the first objective of this paper is to use the hierarchical model architecture to further enhance a state-of-the-art Transformer-based risk prediction model, BEHRT[10], [16]. The hierarchical BEHRT (Hi-BEHRT) model will be able to handle risk prediction for patients with

Y. L., M. M., G. S., S. S., A. H., D. C., and K. R. are with the Deep Medicine, Oxford Martin School, University of Oxford, Oxford, United Kingdom. (Corresponding author: Kazem Rahimi, e-mail: kazem.rahimi@wrh.ox.ac.uk)

T. L. is with the Department of Computer Science, University of Oxford, Oxford, United Kingdom

more comprehensive and longer EHR than BEHRT. We will investigate and compare model performance on incident risk prediction for four diseases: heart failure (HF), diabetes, chronic kidney disease (CKD), and stroke. In addition, by contrast with BEHRT which used diagnosis only as model input, we will also include information about medications, procedures, tests, blood pressure (BP) measurement, drinking status, smoking status, and body mass index (BMI), which are important predictors of outcomes and periodically measured in routine clinical practice.

In addition to the model architecture, the pre-training strategy is also a critical component in modelling. Most of the current pre-training methods for Transformer are either based on the masked language model (MLM) or the sequence pair prediction (e.g., the next sentence and the next segment prediction) or both[9], [10], [12], [17]. Because the definition of the "next sentence" in EHR is not as clearly defined as the concept in natural language processing, MLM is deemed a more suitable approach in EHR applications[9], [10]. In terms of the training task, MLM predicts the masked records in a sequence using their contexts. However, in the hierarchical structure, the lower-level feature extractor has transformed the records into high dimensional representations, thus, defining a clear label for MLM training is difficult. Some of the previous works using hierarchical Transformer only initialized the weight of embeddings or certain components of a model using weights pre-trained on other tasks[15]. Therefore, pre-training the entire hierarchical Transformer architecture and fine-tuning on down-stream tasks using EHR is still not well-established. Recent proposed self-supervised pre-training framework on contractive learning, more specifically, bootstrap your own latent[18] (BYOL) provides an alternative approach for pre-training. It directly compares the latent representations from the model, expecting the different augmentations of the same input to have similar representations. Therefore, it is flexible in terms of the model architecture and can be adopted to the Hi-BEHRT model. However, the reported advantage of BYOL in pre-training and overcoming data scarcity has only been tested on an image dataset and whether this can have some advantages when applied to sequential EHR is unclear. To this end, the second objective of this paper is to evaluate the usability of BYOL for sequential Hi-BEHRT model using EHR.

## II. Methods

### A. Data source and cohort selection

We undertook a cohort study in a large population of primary care patients using Clinical Practice Research Datalink (CPRD)[19]. It is one of the largest deidentified longitudinal EHR dataset that contains patient data from approximate 7% of the UK population[20]. Primary care records from CPRD can also link to secondary care records from Hospital Episodes Statistics Admitted Patient Care data and death registration data from the Office for National Statistics. It is broadly representative in terms of age, sex, and ethnicity. We identified an open cohort of patients aged 16 years and older and contributed to data between Jan 1, 1985 and Sep 30, 2015. Patients were eligible for inclusion if their records were labelled as "accept" by the CPRD quality control[19] and they were linked to Hospital Episodes Statistics.

Among the identified patients, we split them into 60%, 10%, and 30% for training, tuning, and validation for the risk prediction task, respectively. Additionally, the patients in the training and tuning cohorts were also used for pre-training. Within each risk prediction cohort, we identified two important dates for each patient, an incident date of the event of interest and a baseline date. All the records before the baseline date were used as learning period to predict the 5-year risk of an event of interest after it. Given the known inaccuracies in recording of timing of events, we also ignored the events that occurred within 1-year after the baseline date. We achieved this by firstly identifying the incident date for all positive cases, and randomly selected a baseline date within a 1-year to 5-year window before the incident date. For patients who did not have recorded event of interest, we considered them as negative patients, and randomly selected a baseline date for each of them with a guarantee of having at least 5 years of records after the baseline date. For all risk prediction tasks, we also excluded patients who had less than 3 years of records (learning period) before the baseline date.

### B. Case identification

In this study, we focused on 5-year risk estimation of the incidence of HF, diabetes, CKD, and stroke. HF was defined as a composite condition of rheumatic heart failure, hypertensive heart and disease with (congestive) heart failure and renal failure, Ischemic cardiomyopathy, chronic cor pulmonale, congestive heart failure, cardiomyopathy, left ventricular failure, and cardiac, heart, or myocardial failure[21]; Diabetes was defined as a composite condition of type 1 and type 2 diabetes mellitus, malnutrition-related diabetes mellitus, other specified diabetes mellitus, and pre-existing malnutrition-related diabetes mellitus[22]; CKD included chronic kidney disease from stage 1 to stage 5, kidney transplant failure and rejection, obstructive and reflux uropathy, acute renal failure, nephrotic syndrome, hypertensive renal failure, type 1 and 2 diabetes mellitus with kidney complications, chronic tubulo-interstitial nephritis[23]–[25]; and the identification of stroke used a composite condition of cerebrovascular diseases, subarachnoid hemorrhage, intracerebral hemorrhage, sequelae of cerebrovascular disease, cerebral infarction, and occlusion and stenosis of cerebral arteries[23]. We further defined the incident HF, diabetes, CKD, and stroke as the first record of corresponding disease in primary care or hospital admission records from any diagnostic position. The list of International classification of diseases, tenth revision[26] (ICD-10) codes been used to identify these four diseases can be found in Supplementary.

### C. Data processing

We included records from diagnoses, medications, hospital procedures, GP tests, BP measurements (both systolic and diastolic pressure), drinking status, smoking status, and BMI. For diagnosis, we used ICD-10 as the standard format, thus, we mapped all diagnostic records from the primary care (Read[27]) to the ICD-10 level 4 as the work proposed by Hassaine et al.[28]. In brief, all codes were mapped to ICD-10 using a vocabulary provided by NHS digital[29] and SNOMED-CT[30]. The NHS digital vocabulary had higher priority than the SNEMOED-CT wherever a conflict occurred. For medications, procedure, and test, we used British National Formulary (BNF)[31] coding scheme in the section level, The Office of Population Censuses and Surveys (OPCS) Classification of Interventions and Procedures codes, and Read code, respectively. Both drinking and smoking status were recorded as categorical values, including current drinker/smoker, ex, and non. For continues values, we included systolic pressure, diastolic pressure, and BMI within range 80 to 200 mmHg, 50 to 140 mmHg, and 16 to 50 kg/$m^2$, respective. Afterwards, we categorized systolic and diastolic pressure into bins with a 5-mmHg step size (e.g., 80-85 mmHg). BMI was processed the same way with a step size 1 kg/$m^2$. In the end, all records of a patient were formatted as a sequence and ordered by the event date. For the convenience of modelling, additional feature of each record, age, was also calculated by the event date and the patient's date of birth. All the records in the pre-training dataset were used for pre-training. However, in the risk prediction dataset, only records before the baseline date were used as learning period for risk prediction tasks.

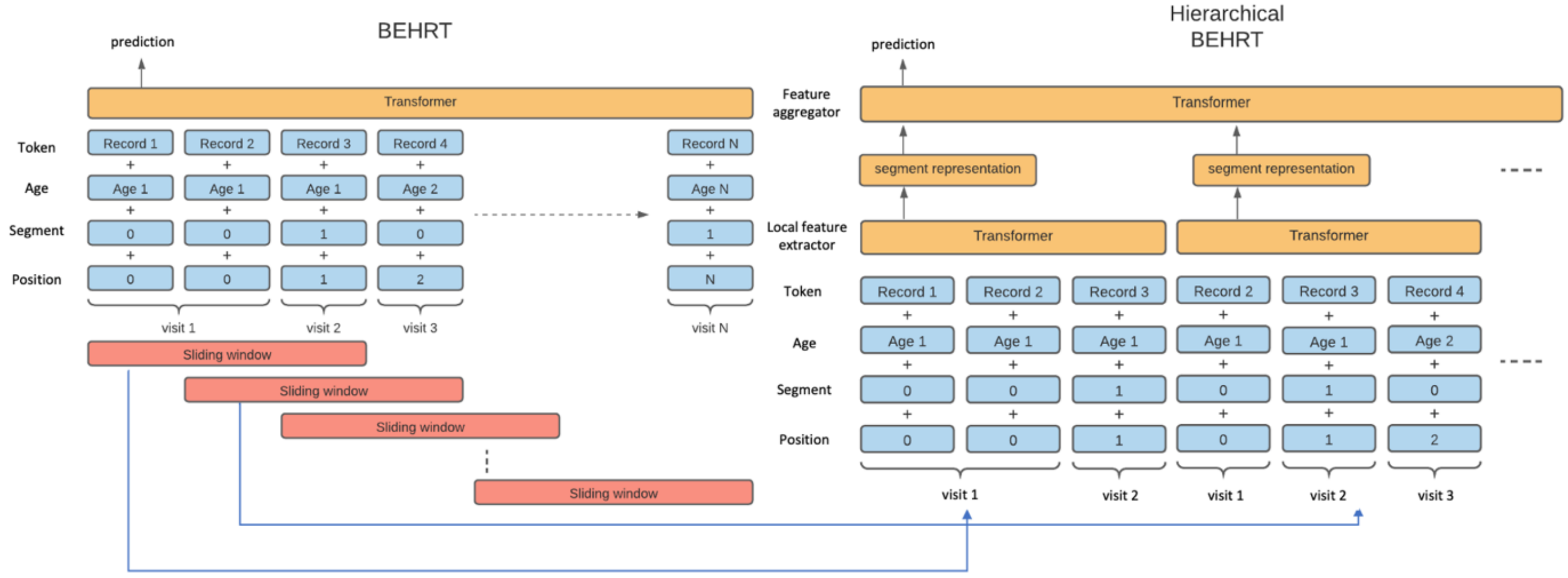

Fig. 1. Model architecture for BEHRT and hierarchical BEHRT. Despite the segment representation appearing non-overlapping (right), they are built on overlapping tokens in the sliding window (left).

## *D. Training and validation of the models*

We used the same hyper-parameters for pre-training and all four risk prediction tasks. The hyper-parameters were tuned on the tuning set of HF risk prediction task. Afterwards, we reported area under the receiver operating characteristic (AUROC) curve and area under the precision recall (AUPRC) curve validated on the validation set of each risk prediction task with model trained on data from both training and tuning set.

## *E. Model derivation and development*

Figure 1 uses a hypothetical patient to show the model architecture of the state-of-the-art BEHRT model[10] and the proposed Hi-BEHRT model. For the BEHRT model, four types of embeddings were taken as inputs. The token embeddings were projected from all available codes or categorical variables of records from diagnosis, medication, procedure, test, BP measurement, drinking status, smoking status, and BMI. The age embeddings were representations of the age in year. The segmentation embeddings alternatively changed between different visits with value 0 and 1 and the position embeddings monotonically increase across different visits. The embedding of each encounter (i.e., a record and its corresponding age, segmentation, and position) is represented by the summation of the record, age, segmentation, and position embeddings. The inputs were followed with a Transformer model to extract the feature interactions, and the latent representation of the first timestep in the last layer was projected by a pooling layer for risk prediction.

Instead of extracting the interaction of records in the entire medical history, we used a similar idea as the work proposed by Pappagari el al.[15] for the Hi-BEHRT. The Hi-BEHRT model used a sliding window to segment the full medical history into smaller segments and applied a Transformer as a local feature extractor to extract temporal interaction within each segment. Because medical records naturally have stronger correlation when they are closer in time, we would expect the local feature extractor learns the most representative latent representation for each segment. Afterwards, we applied another Transformer as a feature aggregator to globally summarize the local features extracted in all segments. Similarly, a risk prediction was made based on the latent representation from a pooling layer.

## *F. Pre-training*

We applied BYOL[18] for the self-supervised pretraining. It was originally designed for image representation learning. In this work, we implemented this idea with MLM[12] and adapted it to pre-train our Hi-BEHRT model.

### *1) BYOL*

The main idea of BYOL is that the different augmentations of the same data should have similar representations. Therefore, this framework has two networks for training: an online network and a target network. The online network includes an encoder, a projector, and a predictor; the target network has the same architecture as the online network but with a different set of weights. As shown in Figure 2, the objective of the task is to minimize the mean squared loss between the output of the online predictor and the output of the target projector. The weights of the online network are updated through backpropagation; however, the weights of the target network are the exponential moving average of the weights of the online network. This can be shown as following,

$$\zeta \leftarrow \tau\zeta + (1-\tau)\theta,$$

where $\zeta$ and $\theta$ are the weights of the target network and the online network, respective, and $\tau$ is a decay factor between 0 and 1.

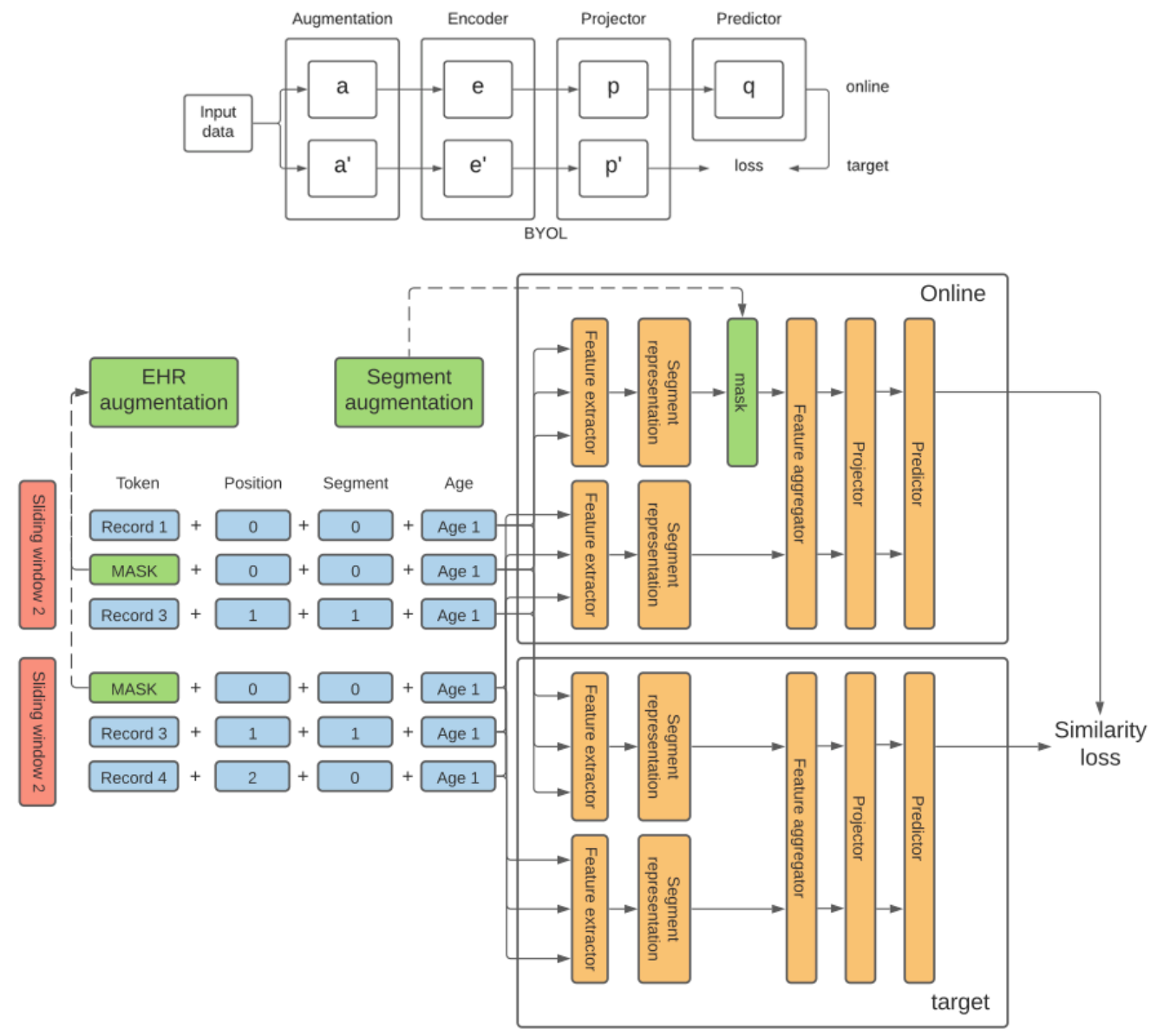

Fig. 2. Illustration of BYOL and BYOL for hierarchical BEHRT. BYOL (top) includes four components, data augmentation, encoder, projector, and predictor. Both online and target network have the same model architecture. BYOL for hierarchical BEHRT (bottom) has similar framework, the segment augmentation corresponds to the augmentation in BYOL (top), and the segment augmentation is conducted on temporal features extracted by local feature extractor.

*2) BYOL for hierarchical BEHRT*

In our study, the encoder was the Hi-BEHRT model, and both the projector and the predictor were a one hidden layer multi-layer perceptron network. Additionally, to adapt BYOL for EHR, we applied two augmentation strategies at present study for the raw EHR and the segment representations, respectively. We referred them as EHR augmentation and segment augmentation as shown in Figure 2. The purpose of EHR augmentation is to enrich the EHR data and increase the data diversity. Segment augmentation is the key component of applying BYOL for Hi-BEHRT. Firstly, we only applied segment augmentation to a certain proportion of time steps in the latent segment representation space, similar to the idea of MLM in BERT[12]. Additionally, unlike the BYOL, which applies augmentation to both online and target network, we only augmented the segment representations in the online network and did not augment anything in the target network (as shown in Figure 2). The intuition is that we expect the context can provide sufficient information for the network to reproduce the representation for the augmented time steps. Therefore, our objective was to use the online and the target network to reduce the dissimilarity of the representation of the augmented time steps and their original representations, and this can be achieved by optimizing the summation of the similarity loss of all augmented time steps.

*3) EHR augmentation*

EHR augmentation was applied before sliding window. We applied random crop for the EHR with a probability $P_c$ and we randomly masked each token with a probability $P_m$. Random crop means randomly select a subset period of EHR.

*4) Segment augmentation*

Segment augmentation was applied to segment representations before feeding them into feature aggregator. We either masked a segment representation by casting it to 0 or added a Gaussian noise to deprecate the latent representation.

## G. Implementation details

*1) Data augmentation*

For pre-training, we firstly used 50% probability ($P_c$) to random crop a EHR sequence, followed by a random mask with 20% probability ($P_m$) to mask each record. We additionally had 50% probability to augment the segment representation, including 85% probability of masking a representation to 0 and 15% probability of adding Gaussian noise to deprecate it.

*2) Architecture*

For comparison, we implemented both BEHRT and Hi-BEHRT model. The BEHRT model used hyper-parameters searched in another work[32], and similar parameters were also used for the Hi-BEHRT model. More specifically, for BEHRT model, we used hidden size 150, number of attention heads 6, intermediate size 108, number of layers 8, maximum sequence length 256, dropout rate 0.2, and attention dropout rate 0.3. For Hi-BEHRT model, we extended the maximum sequence length to 1220, which covered the full EHR for approximate 97% patients across all four risk prediction tasks. We used 50 and 30 as the size and the stride for the sliding window, respectively. Similarly, we used hidden size 150, number of attention heads 6, intermediate size 108, dropout rate 0.2, and attention dropout rate 0.3. The number of layers for the feature extractor and feature aggregator is 4 and 4, respectively. Additionally, we used moving average decay factor 0.996 as recommended by BYOL to update the target network, and hidden size 150 for both projector and predictor in the pre-training stage for the Hi-BEHRT model. Gaussian error linear units (GELU) were used as activation function for both models.

*3) Model complexity*

For a Transformer model, the space and time complexity of the self-attention are $O(L^2 + Ld)$ and $O(L^2 d)$, respectively, where L represents the sequence length and d represents the hidden dimension size. Comparing to BEHRT, the sequence lengths of the local feature extractor and the feature aggregator in the Hi-BEHRT are the sliding window size (50) and the number of segments (39 for maximum sequence length 1220 with window size 50 and stride size 30), respectively. Both of them are substantially shorter than the sequence length of the BEHRT model (256). Even though the space complexity of the local feature extractor increases linearly as the increase of the number of segments, the Hi-BEHRT is still much smaller than the BEHRT model, and the longer the EHR sequence, the bigger the difference.

*4) Optimization*

We used Adam optimizer[33] with a three-stage learning rate schedule[34] and early stopping strategy, over 100 epochs, to train BEHRT and Hi-BEHRT for risk prediction. The three-stage learning schedule included 10%, 40%, and 50% epochs for warm-up, hold, and cosine decay, respectively. We used batch size 128 and swept the hold learning rate among 5e-5, 1e-4, and 5e-4, and reported the one with the best performance for each of the risk prediction task. In terms of the early stopping strategy, we stopped training once the validation loss doesn't decrease for 6 epochs. We followed the same strategy to fine-tune the pre-trained Hi-BEHRT model for the downstream risk prediction tasks. For the pre-training task, we used a similar set up as the BYOL paper[18]. More specifically, we used a stochastic gradient decent optimizer with momentum 0.9, with a cosine decay learning rate schedule, over 1000 epochs, with a warm-up period of 10 epochs. However, due to resource limitation, we were only able to finish 35 epochs for a 10-day training using batch size of 256 split over 2 GPUs.

## III. Results

### A. *Descriptive analysis of the cohorts*

TABLE I
Descriptive analysis of patients

| | HF | Diabetes | CKD | Stroke |
|---|---|---|---|---|
| **General characteristics** | | | | |
| No. of patients | 1,975,630 | 1,885,924 | 1,889,925 | 1,730,828 |
| No. (%) of positive cases | 94,495 (4.7) | 91,659 (4.9) | 175,440 (9.3) | 224,442 (12.9) |
| **Baseline characteristics** | | | | |
| Mean No. of visits per patient | 64.11 | 60.65 | 65.33 | 59.22 |
| Mean No. of codes per visit | 4.44 | 4.32 | 4.44 | 4.45 |
| Mean (SD) learning period (year) | 8.38 (3.99) | 8.31 (3.96) | 8.40 (4.02) | 8,18 (3.85) |
| Mean (SD) age (year) | 53.04 (18.18) | 52.76 (18.33) | 53.89 (18.16) | 52.58 (18.11) |
| % of patients with at least one diagnosis code | 100 | 100 | 100 | 100 |
| % of patients with at least one medication code | 95.9 | 95.9 | 96.2 | 95.1 |
| % of patients with at least one procedure code | 60.9 | 60.5 | 59.5 | 61.6 |
| % of patients with at least one test code | 88.8 | 88.3 | 88.9 | 88.3 |
| % of patients with at least one BP measurement | 88.1 | 87.7 | 88.2 | 87.4 |
| % of patients with at least one BMI measurement | 70.2 | 68.7 | 70.2 | 69.4 |
| % of patients with at least one drinking status | 62.9 | 61.9 | 63.0 | 62.0 |
| % of patients with at least one smoking status | 83.5 | 82.8 | 83.6 | 82.9 |
| Mean No. of diagnosis codes per patient | 18.84 | 18.35 | 18.57 | 17.27 |
| Mean No. of medication codes per patient | 94.70 | 85.83 | 98.79 | 85.73 |
| Mean No. of procedure codes per patient | 4.40 | 4.28 | 4.04 | 4.39 |
| Mean No. of test codes per patient | 80.13 | 72.03 | 80.76 | 75.80 |
| Mean No. of BP measurement per patient | 15.43 | 14.28 | 15.58 | 14.51 |
| Mean No. of BMI measurement per patient | 2.69 | 2.28 | 2.71 | 2.56 |
| Mean No. of drinking status per patient | 1.38 | 1.28 | 1.40 | 1.32 |
| Mean No. of smoking status per patient | 18.84 | 18.35 | 18.57 | 17.27 |
| Median (IQR) EHR length for learning period | 156 (267) | 148 (252) | 158 (285) | 146 (255) |
| No. (%) of patients with EHR length for learning period > 256 | 666,450 (33.7) | 596,072 (31.6) | 649,464 (34.4) | 546,437 (31.5) |

SD: standard deviation, IQR: interquartile range

After cohort selection, we included 2,844,733 patients in the pre-training cohort, and 2,438,352, 406,381, and 1,219,078 patients in the training, tuning, and validation cohort, respectively. Moreover, 1,995 diagnosis codes, 378 medication codes, 275 test codes, 960 procedure codes, 24 systolic BP categories, 17 diastolic BP categories, and 34 BMI categories were considered for modelling. The descriptive analysis of the selected cohort for HF, diabetes, CKD, and stroke risk prediction are shown in Table I. We define the learning period as the time period of EHR up to the baseline. Due to the random selection for the baseline for the negative patients and the exclusion of patients who has less than 3 years of learning period, there are certain variability for the number of patients included across different risk prediction tasks.

### B. *Model performance evaluation*

We accessed the performance of Hi-BEHRT on four risk prediction tasks and compared it with the performance of BEHRT and the Hi-BEHRT finetuned on self-supervised pre-training task on the validation set. Additionally, we applied subgroup analysis to evaluate model performance on patients with different learning period in respect of the EHR length. For patients who have EHR length longer than the pre-defined maximum length in BEHRT model (256), we used the latest 256 EHR records in terms of the event date for modelling and the Hi-BEHRT model is processed the same way but with maximum length 1220. We ran each experiment over three seeds and reported the average performance.

*1) Evaluation on risk prediction*

Table II shows the performance comparison of the BEHRT model and the Hi-BEHRT model. With smaller model size and less model complexity, the Hi-BEHRT model shows superior performance and outperforms the BEHRT model on all risk prediction tasks with 1% to 5% and 3% to 8% absolute improvement for AUROC and AUPRC respectively.

TABLE II
MODEL PERFORMANCE EVALUATION FOR BEHRT AND HI-BEHRT ON RISK PREDICTION.

| Task | BEHRT | | Hi-BEHRT | |
|---|---|---|---|---|
| | AUROC | AUPRC | AUROC | AUPRC |
| HF | 0.93 | 0.71 | 0.96 | 0.77 |
| Diabetes | 0.88 | 0.66 | 0.93 | 0.74 |
| CKD | 0.91 | 0.76 | 0.93 | 0.79 |
| Stroke | 0.89 | 0.76 | 0.90 | 0.79 |

*2) Semi-supervised training for risk prediction*

TABLE III
PERFORMANCE COMPARISON ON TRANSFERRING PRE-TRAINED REPRESENTATIONS TO RISK PREDICTION TASK

| Task | Subset sample size (% of full training set) | No. (%) of positive case | Hi-BEHRT without pre-training | | Hi-BEHRT with pre-training | |
|---|---|---|---|---|---|---|
| | | | ROC | PRC | ROC | PRC |
| HF | 13,827 (1%) | 633 (4.5) | 0.84 | 0.19 | 0.86 | 0.23 |
| | 69,136 (5%) | 3,317 (4.8) | 0.86 | 0.23 | 0.88 | 0.26 |
| Diabetes | 13,201 (1%) | 643 (4.9) | 0.70 | 0.13 | 0.73 | 0.16 |
| | 66,003 (5%) | 3,223 (4.9) | 0.79 | 0.22 | 0.79 | 0.22 |
| CKD | 13,228 (1%) | 1,224 (9.3) | 0.73 | 0.22 | 0.76 | 0.26 |
| | 66,140 (5%) | 6,110 (9.2) | 0.77 | 0.26 | 0.80 | 0.33 |
| Stroke | 12,114 (1%) | 1,579(13.0) | 0.65 | 0.21 | 0.67 | 0.23 |
| | 60,568 (5%) | 7,827(13.0) | 0.68 | 0.24 | 0.76 | 0.41 |

% of positive cases represents the percentage of positive cases in the dataset, for example, HF has 633 positive cases in the 1% subset, and it is around 4.5 percent over 13,827 patients. ROC is AUROC and PRC is AUPRC

Next, we evaluated the performance of the Hi-BEHRT obtained when fine-tuning pre-trained representation on the risk prediction task with a small subset of the training dataset (i.e., both training and tuning sets). We bootstrapped 1% and 5% of the training dataset for three runs and reported the averaged AUROC and AUPRC evaluated on the validation set. Table III shows that comparing to Hi-BEHRT model trained from scratch, the pre-trained representation can provide substantial improvement when fine-tuning on small dataset. In general, for models fine-tuned on the pre-trained representation, they can achieve similar performance as the model trained on 5% of training dataset without pre-training when only been trained on 1% of the training dataset. Additionally, the improvement is clearer when there is a higher percentage of positive cases. For example, CKD with 9.3% of positive cases on 5% subset of training dataset improves 7% and 3% for AUPRC and AUROC, respectively; and stroke with 13% of positive cases on 5% subset of training dataset improves 12% and 7% for AUPRC and AUROC, respectively, comparing to the Hi-BEHRT model without pre-training.

*3) Subgroup analysis for patients with different learning period in respect of EHR length*

To better understand how BEHRT and Hi-BEHRT handle patients with longer learning period, we evaluated model performance on subgroups of patients that have EHR length 0 – 256 and longer than 256 in learning period, respectively. Additionally, because patients with longer EHR (i.e., > 256) have higher percentage of positive cases, we included one more experiment, which preserved all negative samples and bootstrapped a subset of positive samples to create an evaluation dataset with similar proportion of positive cases as the subgroup of patients with shorter EHR (i.e., 0 - 256), for a fairer comparison. We repeated bootstrap for 5 times and reported the averaged results. Here, 256 is the maximum EHR length a BEHRT model can handle in our experiment.

TABLE IV
SUBGROUP ANALYSIS FOR PATIENTS WITH DIFFERENT EHR LENGTH IN LEARNING PERIOD

| Sample size | % of positive cases | EHR length | BEHRT ROC | BEHRT PRC | Hi-BEHRT ROC | Hi-BEHRT PRC |
|---|---|---|---|---|---|---|
| HF | | | | | | |
| 392,161 | 2.3 | 0 - 256 | 0.91 | 0.57 | 0.95 | 0.69 |
| 189,908 | 9.0 | > 256 | 0.91 | 0.75 | 0.94 | 0.81 |
| 176,802 | 2.3 | > 256 | 0.91 | 0.64 | 0.95 | 0.72 |
| Diabetes | | | | | | |
| 386,373 | 3.2 | 0 - 256 | 0.88 | 0.61 | 0.92 | 0.68 |
| 171,777 | 8.2 | > 256 | 0.87 | 0.7 | 0.93 | 0.80 |
| 169,928 | 3.2 | > 256 | 0.86 | 0.63 | 0.93 | 0.74 |
| CKD | | | | | | |
| 371,577 | 5.3 | 0 - 256 | 0.89 | 0.68 | 0.90 | 0.69 |
| 184,597 | 15.7 | > 256 | 0.9 | 0.81 | 0.93 | 0.84 |
| 164,373 | 5.3 | > 256 | 0.90 | 0.72 | 0.93 | 0.75 |
| Stroke | | | | | | |
| 354,698 | 10.3 | 0 - 256 | 0.89 | 0.74 | 0.89 | 0.75 |
| 157,284 | 18.3 | > 256 | 0.88 | 0.79 | 0.92 | 0.84 |
| 143,298 | 10.3 | > 256 | 0.88 | 0.74 | 0.92 | 0.80 |

ROC is AUROC and PRC is AUPRC

Table IV shows that Hi-BEHRT outperforms BEHRT model in all subgroups for all four risk prediction tasks. More specifically, for patients who have EHR length longer than the BEHRT model's capacity (256), Hi-BEHRT model shows approximate 3% - 6% and 3% - 10% improvement in terms of AUROC and AUPRC, respectively. Additionally, when comparing subgroups of patients with EHR length 0 – 256 and > 256 but with similar percentage of positive cases, Hi-BEHRT shows a more substantial improvements with the inclusion of more records, proving its advantage on processing long EHR. Moreover, we also notice for HF risk prediction, the performance of BEHRT is considerably worse than the Hi-BEHRT on short EHR subgroup. Even within the BEHRT model itself, its performance on short EHR subgroup is far worse than its performance on long EHR group. This is quite unusual considering the differences of its performance between long and short EHR subgroups are relatively small for other diseases. One potential reason is HF is an extremely imbalanced task with only 4.7% positive cases and most of the positive cases are within the subgroup of patients who have long EHR. Therefore, BEHRT model can be severely biased and better at identifying positive cases in long EHR group. However, with the inclusion of almost the entire EHR sequence and using sliding window to constrain the receptive field for the local feature extractor, Hi-BEHRT has a better focus on identifying local temporal features and distinguishing patterns between long and short EHR sequence, leading to a more comprehensive risk estimation under both circumstances.

### C. *Ablation analysis*

In this section, we use HF risk prediction task as an example to present ablations on Hi-BEHRT to give an illustration of its behavior and performance. We reported the average performance of models trained over three random seeds.

#### 1) *Training size and performance*

In Table III, we fine-tuned the pre-trained representations of Hi-BEHRT over 1% and 5% of the training dataset. In this section, we used HF risk prediction as an example to further explore the difference of model performance between with and without pre-training over 1%, 5%, 10%, 20%, 50%, and 100% of the training dataset. As shown in Figure 3, model with pre-training has better performance on small dataset (i.e., 1%, 5%, and 10%), and the model with pre-training substantially outperforms the model without pre-pretraining for 31% and 4% in terms of AUPRC and AUROC, respectively, on a subset of training set with 10% of samples. In general, all metrics from both models trained with and without pre-training follow the power law learning curve[35], which includes the small data region (model struggle to learn from a small number of training samples), power-law region (a region that substantially improves model performance with inclusion of more training samples), and irreducible error region (a region that represents the lower-bound error and model will be unable to improve if within this region). The figure shows the model with pre-training can reach the power-law region with smaller sample size. Additionally, figure also shows that our model reaches the irreducible error region with around 50% samples.

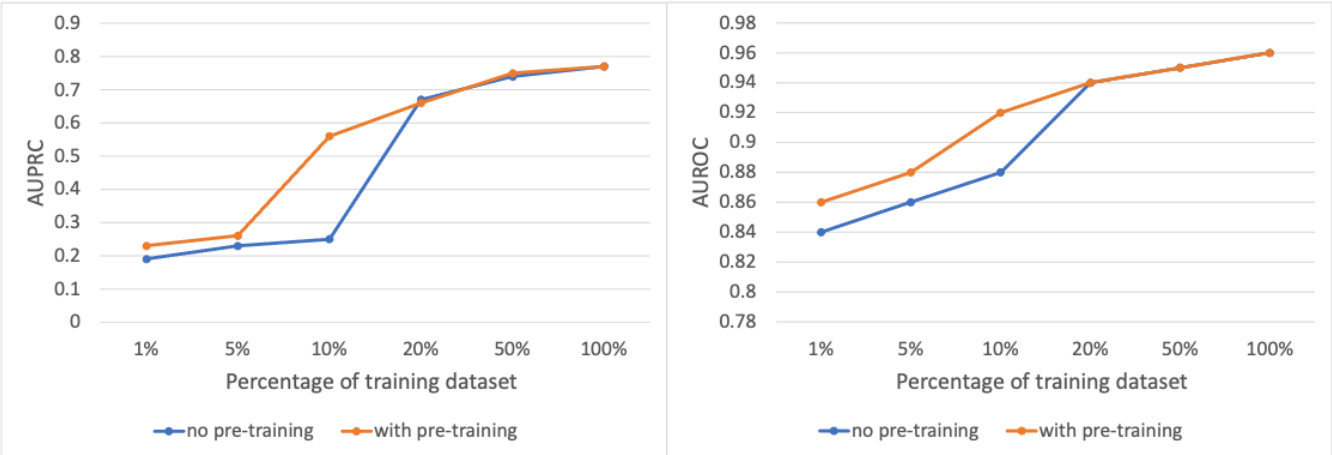

Fig. 3. Performance of Hi-BEHRT trained on a fraction of training dataset

#### 2) *Ablation of modality*

EHR from different modalities can potentially provide different information for modelling. Diagnosis and medication are commonly used modality in risk prediction task. In this experiment, we investigated additional EHR from procedure, test, BP measurement, BMI measurement, drinking and smoking status, and compared how the richness of the EHR affected the model performance. To this end, we used the Hi-BEHRT model reported in Table II and evaluated the model performance on all patients with the inclusion of one additional modality a time besides diagnosis and medication. As shown in Figure 5, we see a trend that with the inclusion of more modalities, the model performance in terms of AUPRC and AUROC improves. However, the contribution of a modality is highly related to the frequency of that modality in the dataset and its importance. Firstly, model greatly improves with the inclusion of test, BP measurement, and smoking status. One of the most obvious reasons is that all of them have relatively high frequency of recording in the dataset (Table I). On the contrary, BMI, drinking status, and procedure have very poor contribution due to their scarcity, and with the inclusion of other modalities, their occurrence and contribution become more negligible. More specifically, for example, there were 960 procedure codes included in the dataset, however, there were only 4.4 procedure related records on average for each patient. Therefore, the scarcity of the procedure codes in the dataset can lead to a biased representation while the model training and potentially limit its contribution to the prediction. Secondly, with similar frequency of recording, the inclusion of BP measurement shows greater improvement than the inclusion of smoking status, indicating BP measurement is a more important modality than smoking status for HF risk prediction.

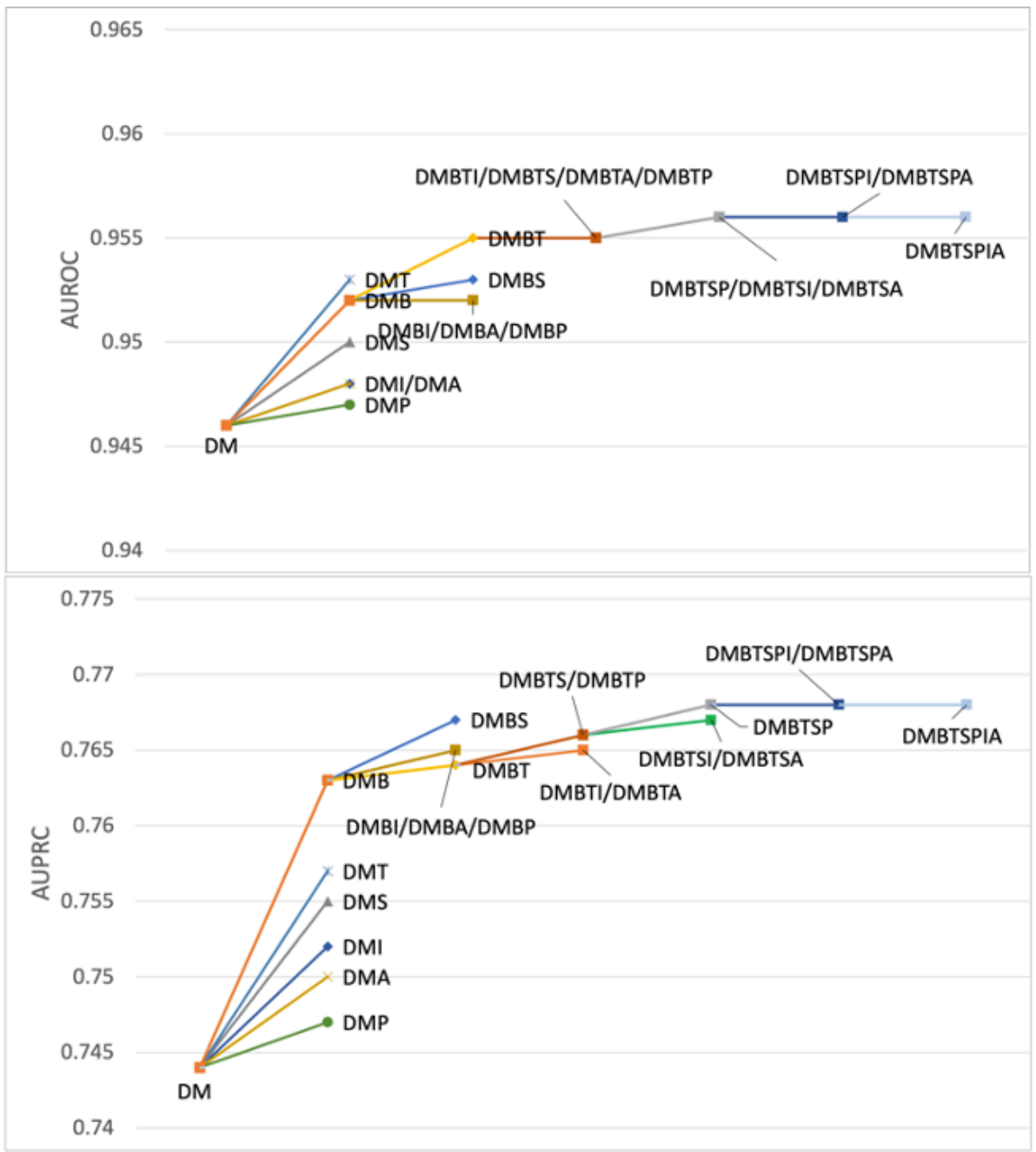

Fig. 4. Ablation on modalities. Evaluate hierarchical BEHRT on patients with inclusion of diagnosis (D), medication (M), procedure (P), test (T), BP measurement (B), BMI (I), smoking status (S), and drinking status (A).

## IV. Discussion

In this paper, we proposed an enhanced BEHRT model, Hi-BEHRT, for risk prediction. It can incorporate long EHR sequences from various modalities, address the shortcomings of vanilla transformers in processing long sequential data, and avoid missing important historical information in risk prediction. With the capability of using the full medical records, Hi-BEHRT outperformed the BEHRT model in terms of AUROC (1%-5%) and AUPRC (3%-8%) for all four investigated (HF, diabetes, CKD, and stroke) risk prediction tasks.

In addition to using diagnoses only in BEHRT, we further investigated EHR from other modalities in this work, including medication, procedure, test, BP measurement, BMI, drinking status, and smoking status. Since they provided additional information for modelling, as expected, the Hi-BEHRT model achieved better performance with the inclusion of more modalities. However, we further observed that the contribution of a modality to the model performance is in general highly related to its frequency of recording in the dataset. It is partially due to the fact that higher frequencies of recording can provide more information for prediction. Another possible explanation is that the information provided by modalities with low frequency are overpowered by the modalities with much higher frequency. This phenomenon has been described in the natural language processing literature for the embedding of words as well which tend to be biased towards higher word frequencies[36]. Therefore, one potential future work will be to investigate how to incorporate features or modalities with low frequency in a more meaningful way for risk prediction.

Furthermore, to better understand the advantage of the Hi-BEHRT comparing to the BEHRT model in terms of handling patients with different learning periods, we further conducted a subgroup analysis to evaluate model performance on patients who have EHR length within the capacity of BEHRT model (i.e., less than or equal to 256) and longer than the BEHRT model's capacity (i.e, more than 256) in the learning period. We found that the Hi-BEHRT model not only substantially outperformed the BEHRT model on risk prediction tasks for patients within the relatively short HER length group, it also greatly improved model performance with the inclusion of more records (i.e., > 256). However, due to the limitation of sequence length in the BEHRT model, the difference of model performance between patients with long EHR and short EHR is very small. Additionally, we notice for very imbalanced outcomes, for example HF, the majority of the positive cases occur in patients with longer EHR length. By making risk prediction with only a fraction of the latest records in BEHRT when patients have long EHR sequence, it treated patients with long EHR the same way as making prediction for patient with short EHR sequence. Therefore, the model can be biased by the positive cases with long EHR records in the training, thus, having relatively poor capability of identifying positive cases with short EHR sequence. On the contrary, with the inclusion of the entire EHR, together with local feature extractor and global feature aggregator to identify temporal and global patterns, the Hi-BEHRT model is more capable of distinguishing different patterns of positive cases in both long and short EHR sequences. Considering the majority (70%) of the population have relatively short EHR length (less than 256) in our risk prediction tasks and probably in most of the cases in reality, this can be an important additional feature of our proposed model.

In addition to model architecture, we also evaluated the usability of a contrastive learning pre-training strategy, BYOL, in this work. We combined the framework, which was originally designed for image pre-training, with the MLM task, and adapted it to pre-train our sequential model. With the pre-training, the Hi-BEHRT model can achieve similar performance using only 1% of training data as the model trained without pre-training using 5% of training data. With additional ablation analysis, we concluded that the pre-training can potentially expanded the power-law region[35] and allowed the model to reach power-law region with smaller data size. However, our results also indicated that the model performance almost saturated when using 50% of training dataset. It means the model achieves the irreducible error region. Future work should investigate more robust model architectures to shift the power-law curve and improve the model accuracy.

One of the major contributions of this work is the provision of a framework for risk prediction with the inclusion of long and comprehensive EHR. With the growing accessibility and usability of EHR systems, risk prediction using long EHR can be inevitable and have important implications for medical practice. To our best knowledge, long sequence modelling has been a challenge, and its application in the context of healthcare and EHR remains unexplored. Our work proposed a potential solution to tackle this problem and investigated its benefit comparing to model that makes prediction using only a fraction of the EHR. Moreover, we provided a self-supervised pre-training framework for the proposed model, and pre-training can adapt risk prediction model to handle tasks with less training data available, which is highly desired in most of the scenarios.

# Supplementary

## VI. Additional information on dataset

In this section, we provide more information on modalities that are not commonly included in the modelling. More specifically, we will introduce procedure and test.

### *A. Procedure*

Procedure is CPRD linked data collected from Hospital Episode Statistics (HES) Admitted Patient Care (EHS APC) data. It is recorded at the point of admission to, or attendances at NHS healthcare providers. All procedure information is coded using the UK Office of Population, Census and Surveys classification (OPCS) 4.6, and procedures that are not covered by OPCS code is not included in the system. Each record in the system is specified with a start and an end date, as well as event date. We used OPCS code and event date to structure the timeline of a patient's EHR history for modelling.

### *B. Test*

Test is recorded in the CPRD test table and coded as Read code. It includes information on history/symptoms, examination/signs, diagnostic procedures, and laboratory procedures. In the experiment, we only used the information in the Read code level, which represents what examinations or procedures are carried out. More detailed quantitative information was excluded.

## VII. Clinical codes used to identify patients with HF, diabetes, CKD, and stroke

TABLE S I
ICD-10 CODES USED TO IDENTIFY PATIENTS WITH HEART FAILURE IN HOSPITAL DISCHARGE RECORDS AND GENERAL PRACTICE RECORDS.

| ICD Code | Description |
|---|---|
| I09.9 | Rheumatic heart failure |
| sI11.0 | Hypertensive heart disease with (congestive) heart failure |
| I13.0 | Hypertensive heart and renal disease with (congestive) heart failure |
| I13.2 | Hypertensive heart and renal disease with both (congestive) heart failure and renal failure |
| I25.5 | Ischemic cardiomyopathy |
| I27.9 | Chronic cor pulmonale |
| I38 | Congestive heart failure due to valvular disease |
| I42.0 | Congestive cardiomyopathy |
| I42.1 | Obstructive hypertrophic cardiomyopathy |
| I42.2 | Nonobstructive hypertrophic cardiomyopathy |
| I42.6 | Alcoholic cardiomyopathy |
| I42.8 | Other cardiomyopathies |
| I42.9 | Cardiomyopathy NOS |
| I50.0 | Congestive heart failure |
| I50.1 | Left ventricular failure |
| I50.2 | Systolic (congestive) heart failure |
| I50.3 | Diastolic (congestive) heart failure |
| I50.8 | Other heart failure |
| I50.9 | Cardiac, heart or myocardial failure NOS |

I38 is mapped from Read code G580400, therefore, the description is slightly different than the description of I38 in the ICD-10.

TABLE S II
ICD-10 CODES USED TO IDENTIFY PATIENTS WITH DIABETES IN HOSPITAL DISCHARGE RECORDS AND GENERAL PRACTICE RECORDS.

| ICD Code | Description |
|---|---|
| E10 | Type 1 diabetes mellitus |
| E11 | Type 2 diabetes mellitus |
| E12 | Malnutrition-related diabetes mellitus |
| E13 | Other specified diabetes mellitus |
| E14 | Unspecified diabetes mellitus |
| O24.2 | Pre-existing malnutrition-related diabetes mellitus |

TABLE S III
ICD-10 CODES USED TO IDENTIFY PATIENTS WITH CKD IN HOSPITAL DISCHARGE RECORDS AND GENERAL PRACTICE RECORDS.

| ICD Code | Description |
|---|---|
| N18.1 | Chronic kidney disease, stage 1 |
| N18.2 | Chronic kidney disease, stage 2 |
| N18.3 | Chronic kidney disease, stage 3 |
| N18.4 | Chronic kidney disease, stage 4 |
| N18.5 | Chronic kidney disease, stage 5 |
| N18.9 | Chronic kidney disease, unspecified |
| T86.1 | Kidney transplant failure and rejection |
| I12.0 | Hypertensive renal failure |
| N00 | Acute nephritic syndrome |
| N03 | Chronic nephritic syndrome |
| N04 | Nephrotic syndrome |
| N05 | Unspecified nephritic syndrome |
| N11 | Chronic tubulo-interstitial nephritis |
| N13 | Obstructive and reflux uropathy |
| N17 | Acute renal failure |
| N19 | Unspecified kidney failure |
| E10.2 | Type 1 diabetes mellitus with kidney complications |
| E11.2 | Type 2 diabetes mellitus with kidney complications |

TABLE S IV
ICD-10 CODES USED TO IDENTIFY PATIENTS WITH STROKE IN HOSPITAL DISCHARGE RECORDS AND GENERAL PRACTICE RECORDS.

| ICD Code | Description |
|---|---|
| I60 | Subarachnoid haemorrhage |
| I61 | Intracerebral haemorrhage |
| I62 | Other nontraumatic intracranial haemorrhage |
| I63 | Cerebral infarction |
| I64 | Stroke, not specified as haemorrhage or infarction |
| I65 | Occlusion and stenosis of precerebral arteries, not resulting in cerebral infarction |
| I66 | Occlusion and stenosis of cerebral arteries, not resulting in cerebral infarction |
| I67 | Other cerebrovascular diseases |
| I68 | Cerebrovascular disorders in diseases classified elsewhere |
| I69 | Sequelae of cerebrovascular disease |
| G45.9 | Transient cerebral ischaemic attack, unspecified |
| G46 | Vascular syndromes of brain in cerebrovascular diseases |

## VIII. Model performance evaluation stratified by baseline age

We evaluated model performance stratified by the baseline age. The comparison was conducted on three subgroups of patients: (1) patients with baseline age between 35 and 50 years old (young adult); (2) patients with baseline age between 50 and 70 years old (middle-aged adult), and (3) patients with baseline age 70-90 years old (older adult). Table S V shows that the hierarchical BEHRT model has better performance across all subgroups, and it substantially outperforms for BEHRT model on HF and diabetes risk prediction tasks, especially for patients with younger age.

TABLE S V
BASELINE AGE STRATIFIED SUBGROUP ANALYSIS

| | | | BEHRT | | Hi-BEHRT | |
|---|---|---|---|---|---|---|
| Sample size | No. (%) of positive cases | Baseline age | ROC | PRC | ROC | PRC |
| HF | | | | | | |
| 154,032 | 1,008 (0.7) | 35-50 | 0.84 | 0.40 | 0.90 | 0.56 |
| 180,416 | 6,878 (3.8) | 50-70 | 0.88 | 0.64 | 0.93 | 0.72 |
| 111,044 | 17,670 (15.9) | 70-90 | 0.86 | 0.75 | 0.90 | 0.80 |
| Diabetes | | | | | | |
| 149,308 | 4,554 (3.1) | 35-50 | 0.87 | 0.60 | 0.92 | 0.69 |
| 167,753 | 12,443 (7.4) | 50-70 | 0.87 | 0.69 | 0.91 | 0.76 |
| 103,866 | 7,932 (7.6) | 70-90 | 0.89 | 0.69 | 0.90 | 0.75 |
| CKD | | | | | | |
| 145,889 | 4,343 (3.0) | 35-50 | 0.88 | 0.62 | 0.89 | 0.64 |
| 176,422 | 13,037 (7.4) | 50-70 | 0.90 | 0.74 | 0.92 | 0.76 |
| 111,727 | 24,875 (22.3) | 70-90 | 0.89 | 0.83 | 0.91 | 0.84 |
| Stroke | | | | | | |
| 136,090 | 11,325 (8.3) | 35-50 | 0.88 | 0.70 | 0.88 | 0.71 |
| 157,789 | 21,392 (13.6) | 50-70 | 0.88 | 0.76 | 0.90 | 0.79 |
| 93,159 | 22,793 (24.5) | 70-90 | 0.87 | 0.82 | 0.89 | 0.84 |

## IX. SIZE AND OVERLAP OF SLIDING WINDOW

For hierarchical BEHRT model, we used sliding window to segment the raw EHR into segments. As shown in Table 4S6 when window size is relatively small (i.e., 50), the size of the stride does not have significant impact in terms of predictive performance, and the bigger stride size can potentially decrease the number of segments and reduce model complexity. However, for the larger window size (i.e., 100), the stride size becomes more important, and some level of overlap between segments is necessary. Without any overlap for window size 100, the AUPRC decreases 4% comparing to the model with stride size 50. Additionally, the analysis shows that not larger window size always the better choice. For instance, AUPRC of window size 100 without overlap decreases 2% comparing to AURPC of window size 50 without overlap. Without overlap, larger window can lead to shorter length in the segment level, and a balance between window size and length of segment might be more preferred in the hierarchical structure.

TABLE S VI
PERFORMANCE OF HF RISK PREDICTION WITH DIFFERENT WINDOW AND STRIDE SIZE

| Window size | stride size | AUROC | AUPRC |
|---|---|---|---|
| 50 | 30 | 0.96 | 0.77 |
| 50 | 50 | 0.95 | 0.76 |
| 100 | 50 | 0.96 | 0.78 |
| 100 | 100 | 0.95 | 0.74 |
| 150 | 150 | 0.95 | 0.74 |

## X. HYPER-PARAMETER TUNING

We set up hierarchical BEHRT with similar hyper-parameters as the BEHRT model and used it as a reference model to tune the hidden size and intermediate size of the Transformer. All experiments were conducted on the 5-year HF risk prediction task. Table S7 shows that hidden size 150 and intermediate size 108 can achieve similar performance as the model with larger size.

TABLE S VII
HI-BEHRT HYPER-PARAMETER TUNING

| Hidden size | Intermediate size | AUROC | AUPRC |
|---|---|---|---|
| 150 | 108 | 0.96 | 0.77 |
| 90 | 108 | 0.95 | 0.74 |
| 240 | 108 | 0.96 | 0.77 |
| 150 | 256 | 0.96 | 0.77 |